\documentclass{Interspeech2024}

\interspeechcameraready

\usepackage{amsmath,graphicx}
\usepackage{hyperref}
\usepackage{booktabs} %
\usepackage{subcaption}
\usepackage{xcolor, colortbl}
\usepackage{xspace}
\usepackage{multirow}
\usepackage{xurl}
\usepackage{tikz}
\usepackage{pgfplots}
\pgfplotsset{compat=1.3}

\usepackage{amsfonts}

\usepackage[
    backend=biber,
    style=ieee,
    citestyle=numeric-comp,
    maxbibnames=4,
    minbibnames=3,
    maxcitenames=3,
    doi=false,isbn=false,url=false,eprint=false
]{biblatex}

\defbibheading{bibliography}[\refname]{}

\title{OWSM v3.1: Better and Faster Open Whisper-Style Speech Models based on E-Branchformer}

\name[affiliation={1}]{Yifan}{Peng}
\name[affiliation={1}]{Jinchuan}{Tian}
\name[affiliation={1}]{William}{Chen}
\name[affiliation={1}]{Siddhant}{Arora}
\name[affiliation={1}]{Brian}{Yan}
\name[affiliation={2}]{Yui}{Sudo}
\name[affiliation={2}]{Muhammad}{Shakeel}
\name[affiliation={1}]{Kwanghee}{Choi}
\name[affiliation={1}]{Jiatong}{Shi}
\name[affiliation={1}]{Xuankai}{Chang}
\name[affiliation={1}]{Jee-weon}{Jung}
\name[affiliation={1}]{Shinji}{Watanabe}

\address{
  $^1$Carnegie Mellon University, USA ~~~~
  $^2$Honda Research Institute Japan, Japan}
\email{yifanpen@andrew.cmu.edu, swatanab@andrew.cmu.edu}


\begin{document}

\maketitle

\begin{abstract}
    Recent studies have highlighted the importance of fully open foundation models. The Open Whisper-style Speech Model (OWSM) is an initial step towards reproducing OpenAI Whisper using public data and open-source toolkits. However, previous versions of OWSM (v1 to v3) are still based on standard Transformer, which might lead to inferior performance compared to state-of-the-art speech encoder architectures. This work aims to improve the performance and efficiency of OWSM without additional data. We present a series of E-Branchformer-based models named OWSM v3.1, ranging from 100M to 1B parameters. OWSM v3.1 outperforms its predecessor, OWSM v3, in most evaluation benchmarks, while showing an improved inference speed of up to 25\%. We further reveal the emergent ability of OWSM v3.1 in zero-shot contextual biasing speech recognition. We also provide a model trained on a subset of data with low license restrictions. We will publicly release the code, pre-trained models, and training logs.\footnote{\url{https://www.wavlab.org/activities/2024/owsm/}}
\end{abstract}

\section{Introduction}

Large speech foundation models have gained popularity recently. Owing to the scaling of model and data sizes as well as the knowledge sharing across languages and tasks, these massively multilingual and multitasking models achieve state-of-the-art (SOTA) performance in various speech processing tasks~\cite{whisper, google-usm, seamless}. OpenAI Whisper~\cite{whisper} is one of the most widely used speech foundation models, which releases pre-trained model weights at five scales from 39M to 1.5B parameters. However, the full development pipeline, including the training data details and model learning dynamics, is unavailable to the public, which could lead to data leakage and concerns about fairness and bias. Recent studies have advocated for open-source reproduction of foundation models, including large language models (LLMs)~\cite{touvron2023llama, liu2023llm360, olmo}, self-supervised speech models~\cite{chen23l_interspeech, wavelablm}, and Whisper-style speech models~\cite{asru23-owsm}. 

The Open Whisper-style Speech Model (OWSM)~\cite{asru23-owsm} is an initial step towards reproducing Whisper-style training using public datasets and an open-source toolkit ESPnet~\cite{espnet}. It supports multilingual automatic speech recognition (ASR), any-to-any speech translation (ST), language identification (LID), and utterance-level alignment. It also publicly releases all scripts, pre-trained model weights, and training logs. To match the design of OpenAI Whisper, the three versions in~\cite{asru23-owsm}, OWSM v1, v2, and v3, adopt the standard Transformer~\cite{transformer} architecture. However, it can lead to suboptimal performance compared to more advanced encoders such as Conformer~\cite{conformer}, Branchformer~\cite{branchformer}, and E-Branchformer~\cite{ebf}.

\begin{figure}[t]
     \begin{tikzpicture}
	\begin{axis}[
		xlabel=Speed-up ($\rightarrow$),
		ylabel=WER ($\leftarrow$),
		xtick={0, 1, 2, 3, 4},
            ytick={8, 10, ..., 14},
    	xmin=0.7,
            xmax=4,
            ymin=6,
            ymax=13.5,
            ylabel shift = -3pt,
            xlabel shift = -4pt,
            label style={font=\footnotesize},
            ticklabel style={font=\scriptsize},
            width=\linewidth,
            height=0.4\linewidth,
	    every axis plot/.append style={thick},
            legend cell align={left},
            legend columns=1,
            legend style={at={(1,0)},anchor=south east,nodes={scale=0.68, transform shape}}
		]
        \addplot[color=cyan, solid, mark=square*, mark options={scale=0.8, solid}] coordinates {
(0.94, 8.1)
(1.81, 9.4)
(2.97, 12.8)
	};
        \addlegendentry{Whisper};
        \node [below] at (axis cs:  0.94, 8.1) {\textcolor{cyan}{\scriptsize medium}};
        \node [above] at (axis cs:  1.81, 9.4) {\textcolor{cyan}{\scriptsize small}};
        \node [below] at (axis cs:  2.97, 12.8) {\textcolor{cyan}{\scriptsize base}};

        \addplot[color=orange, solid, mark=triangle*, mark options={scale=0.8, solid}] coordinates {
(1, 9.6)
	};
        \addlegendentry{OWSM v3};
        \node [above] at (axis cs:  1, 9.6) {\textcolor{orange}{\scriptsize medium}};

        \addplot[color=red, solid, mark=*, mark options={scale=0.8, solid}] coordinates {
(1.24, 7.7)
(2.21, 8.5)
(3.67, 12.1)
	};
        \addlegendentry{OWSM v3.1 (ours)};
        \node [below right] at (axis cs:  1.24, 7.7) {\textcolor{red}{\scriptsize medium}};
        \node [below] at (axis cs:  2.21, 8.5) {\textcolor{red}{\scriptsize small}};
        \node [below] at (axis cs:  3.67, 12.1) {\textcolor{red}{\scriptsize base}};
	
	\end{axis}
\end{tikzpicture}
     \vskip -0.18in
     \caption{
     WER ($\downarrow$) vs. speed-up ($\uparrow$) for English ASR.
     }
     \vskip -0.2in
     \label{fig:en-asr}
\end{figure}
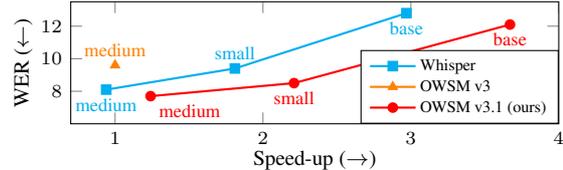

In this work, our goal is to improve the performance and efficiency of the previous OWSM v3 using the same amount of training data (see \autoref{fig:en-asr} for English ASR results). We conduct preliminary experiments to compare Transformer, Conformer, and E-Branchformer encoders and select E-Branchformer due to its faster convergence. We then present new OWSM v3.1 models at three scales: base (101M), small (367M), and medium (1.02B). To stabilize the training of large E-Branchformer models, we propose a piecewise-linear learning rate schedule. Results on extensive benchmarks show that OWSM v3.1 outperforms the previous OWSM v3 in 8 of 9 English ASR, 10 of 11 multilingual ASR, 13 of 19 ST, and 3 of 4 SLUE-PERB~\cite{slue-perb} test sets. Additionally, OWSM v3.1 is 24\% faster for English ASR and 16\% to 25\% faster for ST during inference, owing to the smaller decoder. \autoref{fig:en-asr} shows that OWSM v3.1 even achieves a better trade-off between performance and efficiency than Whisper.
Furthermore, we reveal that OWSM v3.1 has the emergent ability in zero-shot contextual biasing ASR.
To extend the accessibility of our model, we provide a small-sized model trained on a subset of data with low restrictions.
We will publicly release the code, pre-trained models, and training logs to promote transparency and open science.

\section{OWSM v3.1}

\subsection{Model architecture}

Whisper~\cite{whisper} and OWSM v3~\cite{asru23-owsm} adopt the Transformer encoder-decoder architecture~\cite{transformer}. More advanced speech encoders such as Conformer~\cite{conformer} and Branchformer~\cite{branchformer, ebf} have achieved superior results in various speech processing tasks~\cite{conformer-vs-transformer, ebf-vs-conformer}. It is thus natural and promising to explore them in large speech foundation models.
In this work, we demonstrate the effectiveness and scalability of E-Branchformer~\cite{ebf} up to a scale of 1B parameters. 
E-Branchformer is an enhanced Branchformer~\cite{branchformer}, which utilizes parallel branches to capture local and global information and merges them with convolutions. In Whisper-style training, the input audio has a fixed length of 30s, so we simply use the sinusoidal absolute positional encoding.
\autoref{tab:overall-configs} summarizes the model configurations. The proposed OWSM v3.1 mostly follows the design of OWSM v3, except for the encoder. We modify the hidden size and the number of layers to adjust the size of the model. We provide three variants to investigate the scaling behavior, including base (101M), small (367M), and medium (1.02B). Although slightly larger than OWSM v3 and Whisper at the same scale, OWSM v3.1 models exhibit faster inference speeds (see \autoref{fig:en-asr}, \autoref{tab:translation-x-en}, and \autoref{tab:translation-en-x}), mainly due to the smaller decoder.

\begingroup
\setlength{\tabcolsep}{1pt}
\begin{table}[t]
  \caption{Model architectures and training setups. LR (low restriction) is a small-sized model trained on a subset of data with low license restrictions.}
  \label{tab:overall-configs}
  \vskip -0.12in
  \centering
  \resizebox {\linewidth} {!} {
  \begin{tabular}{lcccccccc}
    \toprule
     & \multicolumn{3}{c}{Whisper~\cite{whisper}} & OWSM v3~\cite{asru23-owsm} & \multicolumn{4}{c}{OWSM v3.1 (ours)}\\
     \cmidrule(lr){2-4}
     \cmidrule(lr){5-5}
     \cmidrule(lr){6-9}
     & base & small & medium & medium & base & small & medium & LR\\
    \midrule
    \multicolumn{8}{l}{\textbf{Model architectures}}\\
    Params & 74M & 244M & 769M & 889M & 101M & 367M & 1.02B & 367M\\
    Encoder & \multicolumn{3}{c}{Transformer} & Transformer & \multicolumn{4}{c}{E-Branchformer} \\
    Decoder & \multicolumn{3}{c}{Transformer} & Transformer & \multicolumn{4}{c}{Transformer} \\
    Layers & 6 & 12 & 24 & 24 & 6 & 9 & 18 & 9\\
    Hidden & 512 & 768 & 1024 & 1024 & 384 & 768 & 1024 & 768\\
    Heads & 8 & 12 & 16 & 16 & 6 & 12 & 16 & 12\\
    \midrule
    \multicolumn{8}{l}{\textbf{Training setups}}\\
    Data (h) & \multicolumn{3}{c}{680K} & 180K & \multicolumn{3}{c}{180K} & 70K \\
    Languages & \multicolumn{3}{c}{99} & 151 & \multicolumn{3}{c}{151} & 143\\
    GPU hours & \multicolumn{3}{c}{unknown} & 30.7K & 2.3K & 3.2K & 24.6K & 3.2K\\
    Max LR & 1e-3 & 5e-4 & 2.5e-4 & 2.5e-4 & 1e-3 & 5e-4 & 2e-4 & 5e-4\\
    \bottomrule
  \end{tabular}
  }
  \vskip -0.2in
\end{table}
\endgroup

\subsection{Data preparation}
\label{subsec:dataprep}

We prepare training data using scripts publicly released by~\cite{asru23-owsm}. \autoref{tab:overall-configs} shows the amount of data and the number of languages.
Please refer to \cite{asru23-owsm} for more details. 
We perform the following preprocessing to make the text transcripts more consistent, which affects only a very small amount of data.
\begin{itemize}
    \item We exclude WSJ from the training data due to its different speaking and annotation styles, in which the punctuation is explicitly uttered and annotated as a word.
    \item AMI~\cite{ami-corpus} and VoxForge~\cite{voxforge} provide uppercase transcripts. We convert them to lowercase. Other data remain unchanged.
    \item We merge two language codes ``cmn'' and ``zho'' into ``zho''.
\end{itemize}

Our base, small, and medium models are trained on all 180K hours of data.
To extend the accessibility of our model, we also train a small-sized model using a subset of data with low restrictions (LR): AMI (CC-BY-4.0)~\cite{ami-corpus}, CommonVoice (CC0-1.0)~\cite{commonvoice}, FLEURS (CC-BY-4.0)~\cite{FLEURS}, KsponSpeech (MIT)~\cite{ksponspeech}, LibriSpeech (CC-BY-4.0)~\cite{librispeech-corpus}, Multilingual LibriSpeech (CC-BY-4.0)~\cite{pratap2020mls}, and VCTK (CC-BY-4.0)~\cite{vctk}. This subset contains 70K hours of ASR data but no ST data.

\subsection{Training setups}

Our models are implemented in ESPnet~\cite{espnet} with PyTorch~\cite{pytorch}. We use FlashAttention~\cite{flashattn} to improve training efficiency. The batch size is 256. Our base, small, and medium models are trained for approximately 3 entire passes of the 180K hours of data using 16, 16, and 64 NVIDIA A100 GPUs (40GB), respectively. The low-restriction model follows the setup of OWSM v3.1 small, but uses only 70K hours of data.
\autoref{tab:overall-configs} shows the estimated GPU hours, assuming a stable GPU cluster.

We find it difficult to train models on massively multilingual, multitasking, and long-form speech data.\footnote{Based on our experience, this is mainly due to the 30s long-form data format. Even small models have a hard time converging.} 
A typical strategy to improve convergence is to use a very small learning rate at the beginning of training. However, with the linear warmup schedule, we have to greatly reduce the peak learning rate or increase the warmup steps, both leading to inferior performance according to our preliminary explorations. To alleviate this issue, we propose a piecewise-linear warmup schedule that slowly increases the learning rate at the beginning and more quickly later. Specifically, the learning rate is linearly increased to a very small value (e.g., 5e-5) in the first 30K steps and then linearly increased to the peak learning rate in another 30K steps. After warmup, it is decreased exponentially in the same way as the vanilla version. The proposed piecewise-linear schedule enables successful training of OWSM v3.1.

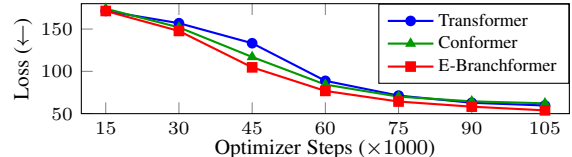
\begin{figure}[t]
     \begin{tikzpicture}
	\begin{axis}[
		xlabel=Optimizer Steps ($\times 1000$),
		ylabel=Loss ($\leftarrow$),
		xtick={0,15,30,...,120},
            ytick={0,50,...,200},
    	xmin=10,
            xmax=110,
            ymin=50,
            ymax=180,
            ylabel shift = -3pt,
            xlabel shift = -4pt,
            label style={font=\footnotesize},
            ticklabel style={font=\scriptsize},
            width=\linewidth,
            height=0.38\linewidth,
	    every axis plot/.append style={thick},
            legend cell align={left},
            legend columns=1,
            legend style={at={(1,1)},anchor=north east,nodes={scale=0.75, transform shape}}
		]

        \addplot[color=blue, solid, mark=*, mark options={scale=0.8, solid}] coordinates {
(15,171.35906982421900)
(30,156.92840576171900)
(45,133.24842834472700)
(60,88.78694915771480)
(75,71.34439086914060)
(90,62.870697021484400)
(105,59.63855743408200)
	}; 
        \addlegendentry{Transformer};

	\addplot[color=green!60!black, solid, mark=triangle*, mark options={scale=0.8, solid}] coordinates {
(15,174.02554321289100)
(30,151.99972534179700)
(45,116.98265838623000)
(60,84.33919525146480)
(75,70.12748718261720)
(90,64.53068542480470)
(105,62.201019287109400)
	}; 
        \addlegendentry{Conformer};

        \addplot[color=red, solid, mark=square*, mark options={scale=0.8, solid}] coordinates {
(15,171.3388671875)
(30,147.723388671875)
(45,104.7358627319340)
(60,76.84737396240230)
(75,64.30705261230470)
(90,58.245155334472700)
(105,53.875972747802700)
	}; 
        \addlegendentry{E-Branchformer};

	\end{axis}
\end{tikzpicture}
     \vskip -0.17in
     \caption{
     Validation loss curves of three encoders.
     }
     \vskip -0.12in
     \label{fig:comparison-encoders}
\end{figure}

\begingroup
\setlength{\tabcolsep}{1pt}
\begin{table}[t]
  \caption{WER ($\downarrow$) of English ASR. 
  \textbf{Bold}: the best result. \underline{Underlined}: OWSM v3.1 outperforms OWSM v3. CV: CommonVoice. LS: LibriSpeech. MLS: Multilingual LibriSpeech.}
  \label{tab:en-asr}
  \vskip -0.1in
  \centering
  \resizebox {\linewidth} {!} {
  \begin{tabular}{ccccccccc}
    \toprule
    \multirow{2}{*}{Test set} & \multicolumn{3}{c}{Whisper} & OWSM v3 & \multicolumn{4}{c}{OWSM v3.1 (ours)}\\
    \cmidrule(lr){2-4}
    \cmidrule(lr){5-5}
    \cmidrule(lr){6-9}
    & base & small & medium & medium & base & small & medium & LR \\
    \midrule
    CV~\cite{commonvoice} & 25.2 & 15.7 & \textbf{11.9} & 14.5 & 21.5 & 14.3 & \underline{12.6} & 12.3\\
    FLEURS~\cite{FLEURS} & 12.4 & 9.6 & \textbf{6.4} & 10.9 & 14.8 & 10.3 & \underline{9.0} & 10.8\\
    LS clean~\cite{librispeech-corpus} & 5.1 & 3.3 & 2.8 & 2.7 & 3.6 & 2.5 & \underline{2.4} & \textbf{2.1}\\
    LS other~\cite{librispeech-corpus} & 12.0 & 7.7 & 6.5 & 6.0 & 9.1 & 5.8 & \underline{\textbf{5.0}} & 5.2\\
    MLS~\cite{pratap2020mls} & 13.4 & 9.1 & 10.2 & 7.4 & 12.0 & 8.1 & \underline{7.1} & \textbf{7.0}\\
    SWBD~\cite{swbd-corpus} & 25.7 & 22.2 & 19.4 & 17.2 & 22.9 & 17.4 & \underline{\textbf{16.3}} & 31.5\\
    TEDLIUM~\cite{tedlium3} & 6.3 & \textbf{4.6} & 5.1 & 4.8 & 7.8 & 5.0 & 5.1 & 9.2\\
    VoxPopuli~\cite{voxpopuli} & 10.2 & 8.5 & \textbf{7.6} & 9.2 & 12.0 & 9.1 &\underline{8.4} & 13.8\\
    WSJ~\cite{wsj} & 5.0 & 4.3 & \textbf{2.9} & 13.4 & 5.3 & 3.8 &\underline{3.5} & 4.9\\
    \midrule
    Ave. WER ($\downarrow$) & 12.8 & 9.4 & 8.1 & 9.6 & 12.1 & 8.5 & \underline{\textbf{7.7}} & 10.8\\
    Speed-up ($\uparrow$) & 2.97x & 1.81x & 0.94x & 1.00x & \textbf{3.67x} & 2.21x & \underline{1.24x} & 2.50x\\
    \bottomrule
  \end{tabular}
  }
  \vskip -0.18in
\end{table}
\endgroup

\begingroup
\setlength{\tabcolsep}{4pt}
\begin{table*}[tb]
  \caption{WER/CER ($\downarrow$) of multilingual ASR. Training data sizes (in hours) are also shown. OWSM v3.1 uses the same amount of training data as OWSM v3. \textbf{Bold}: the best result. \underline{Underlined}: OWSM v3.1 outperforms OWSM v3.}
  \label{tab:multilingual-asr}
  \vskip -0.12in
  \centering
  \resizebox {0.75\linewidth} {!} {
  \begin{tabular}{cccccccccccc}
    \toprule
    \multirow{2}{*}{Test set} & \multirow{2}{*}{Language} & \multirow{2}{*}{Metric} & \multicolumn{4}{c}{Whisper} & \multicolumn{2}{c}{OWSM v3} & \multicolumn{3}{c}{OWSM v3.1 (ours)}\\
    \cmidrule(lr){4-7}
    \cmidrule(lr){8-9}
    \cmidrule(lr){10-12}
    & & & data & base & small & medium & data & medium & base & small & medium\\
    \midrule
    \multirow{8}{*}{MLS~\cite{pratap2020mls}}
    & Spanish & \multirow{8}{*}{WER} & 11.1K & 14.5 & 9.1 & \textbf{6.1} & 2.0K & 11.7 & 18.5 & 10.8 & \underline{9.0} \\
    & French & & 9.8K & 25.2 & 13.6 & \textbf{9.7} & 2.5K & 14.1  & 24.2 & 14.1 & \underline{12.1}\\
    & German & & 13.3K & 19.9 & 11.5 & \textbf{8.1} & 3.7K & 11.9  & 18.7 & 12.4 & \underline{10.8}\\
    & Dutch & & 2.1K & 30.9 & 18.2 & \textbf{12.2} & 1.7K & 17.7 & 28.6 & 19.7 & 18.1\\
    & Italian & & 2.6K & 32.9 & 21.3 & \textbf{15.6} & 0.7K & 24.5 & 33.7 & 21.8 & \underline{20.2}\\
    & Portuguese & & 8.6K & 23.5 & 13.8 & \textbf{8.9} & 0.3K & 28.2 & 44.9 & 26.7 & \underline{21.6}\\
    & Polish & & 4.3K & 25.2 & 12.5 & \textbf{6.8} & 0.3K & 37.0 & 49.7 & 28.5 & \underline{25.2}\\
    \midrule
    AISHELL-1~\cite{aishell-corpus} & Chinese & \multirow{4}{*}{CER} & 23.4K & 39.1 & 25.1 & 15.7 & 16.0K & 7.1 & 12.2 & 7.5 & \underline{\textbf{6.4}}\\
    KsponSpeech clean~\cite{ksponspeech} & \multirow{2}{*}{Korean} & & \multirow{2}{*}{8.0K} & 27.0 & 24.0 & 17.6 & \multirow{2}{*}{1.0K} & 20.5  & 23.8 & 17.2 & \underline{\textbf{16.7}}\\
    KsponSpeech other~\cite{ksponspeech} & & & & 22.9 & 15.4 & \textbf{12.8} & & 22.6 & 26.1 & 18.9 & \underline{18.9}\\
    ReazonSpeech~\cite{reazonspeech} & Japanese & & 7.1K & 54.1 & 32.5 & 25.3 & 18.9K & 11.3 & 11.2 & 8.5 & \underline{\textbf{7.9}}\\
    \midrule
    \multicolumn{3}{l}{Average WER/CER ($\downarrow$)} & - &  28.7 & 17.9 & \textbf{12.6} & - & 18.8 & 26.5 & 16.9 & \underline{15.2} \\
    \bottomrule
  \end{tabular}
  }
  \vskip -0.2in
\end{table*}
\endgroup

\begingroup
\setlength{\tabcolsep}{1.5pt}
\begin{table}[tb]
  \caption{BLEU ($\uparrow$) of X-to-En ST on CoVoST-2~\cite{covost2}. Training data sizes (in hours) are also shown. OWSM v3.1 uses the same amount of training data as OWSM v3. \textbf{Bold}: the best result. \underline{Underlined}: OWSM v3.1 outperforms OWSM v3.}
  \label{tab:translation-x-en}
  \vskip -0.12in
  \centering
  \resizebox {\linewidth} {!} {
  \begin{tabular}{cccccccccccc}
    \toprule
    \multirow{2}{*}{Source} & \multicolumn{4}{c}{Whisper} & \multicolumn{2}{c}{OWSM v3} & \multicolumn{3}{c}{OWSM v3.1 (ours)} \\
    \cmidrule(lr){2-5}
    \cmidrule(lr){6-7}
    \cmidrule(lr){8-10}
    & data & base & small & medium & data & medium & base & small & medium\\
    \midrule
    German & 4.3K & 11.4 & 25.0 & \textbf{33.6} & 0.2K & 16.2 & 7.3 & 15.1 & \underline{17.1}\\
    Spanish & 6.7K & 19.2 & 32.8 & \textbf{39.7} & 0.1K & 20.5 & 10.0 & 19.3 & \underline{22.3}\\
    French & 4.5K & 13.1 & 26.4 & \textbf{34.4} & 0.3K & 21.7 & 11.1 & 20.3 & \underline{22.7}\\
    Catalan & 0.2K & 9.7 & 21.7 & \textbf{29.2} & 0.1K & 16.8 & 9.0 & 16.2 & \underline{18.4}\\
    \midrule
    \multicolumn{2}{l}{Ave. BLEU ($\uparrow$)} & 13.4 & 26.5 & \textbf{34.2} & - & 18.8 & 9.4 & 17.7 & \underline{20.1}\\
    \multicolumn{2}{l}{Speed-up ($\uparrow$)} & 2.14x & 1.80x & 0.98x & - & 1.00x & \textbf{3.23x} & 2.26x & \underline{1.16x}\\
    \bottomrule
  \end{tabular}
  }
  \vskip -0.18in
\end{table}
\endgroup

\section{Experiments}

\subsection{Comparison of encoder architectures}

We first compare different encoders by training small-sized models on 10\% of the training data. These models use the same decoder but different encoders: Transformer, Conformer, or E-Branchformer. Their overall model sizes are kept the same to ensure a fair comparison (366M, 367M, and 367M, respectively). \autoref{fig:comparison-encoders} shows the validation losses within the first 105K steps.\footnote{It takes more than a week for the model to fully converge with 16 GPUs. Due to budget and time limits, we only compare their convergence speeds based on the first 105K steps.}
E-Branchformer converges faster than the others, which is consistent with prior work~\cite{ebf-vs-conformer}. Hence, we adopt E-Branchformer in our main experiments.

\subsection{English speech recognition}
\label{subsec:en-asr}

\autoref{tab:en-asr} shows English ASR results. \autoref{fig:en-asr} visualizes the average word error rate (WER) versus speed-up measured on an NVIDIA A40 GPU. We follow \cite{asru23-owsm} to perform greedy search and apply the Whisper text normalizer before scoring. We have the following observations: (1) Compared to the previous OWSM v3, the proposed OWSM v3.1 medium model performs better in 8 of 9 test sets. The improvement is especially large in CommonVoice, FLEURS, LibriSpeech, Switchboard, VoxPopuli, and WSJ.\footnote{As discussed in \cite{asru23-owsm}, the WSJ training data is used by OWSM v3, but its transcripts are fully uppercased. The model might treat it as another low-resource language, which leads to poor results. In v3.1, we exclude WSJ during training and achieve a significantly lower WER.} This verifies the effectiveness of our E-Branchformer encoder. (2) OWSM v3.1 even achieves lower average WERs than Whisper at each scale, demonstrating its competitive performance, although trained on much less English ASR data (73K vs. 438K hours). (3) OWSM v3.1 is faster during inference than the others at the same scale, primarily due to the smaller decoder. (4) Our small-sized low-restriction (LR) model achieves reasonable performance considering that it is trained on a subset of data (see Section~\ref{subsec:dataprep}).

\subsection{Multilingual speech recognition}

\autoref{tab:multilingual-asr} presents multilingual ASR results. We perform greedy decoding and apply the Whisper text normalizer before calculating word or character error rates (WER/CER). We observe that OWSM v3.1 medium outperforms OWSM v3 in 10 of 11 test sets in various languages, usually by a large margin. Specifically, the average error rate is reduced from 18.8\% to 15.2\%.
Compared to Whisper, OWSM v3.1 still falls behind in many European languages due to limited training data. In contrast, when the data are sufficient (e.g. Chinese and Japanese), OWSM v3.1 achieves strong performance and outperforms Whisper. This reveals the importance of the quantity of training data. In the future, we will include more data from public sources like YODAS~\cite{yodas} to further improve OWSM.

\begingroup
\setlength{\tabcolsep}{4pt}
\begin{table}[tb]
  \caption{BLEU ($\uparrow$) of En-to-X ST on CoVoST-2~\cite{covost2}. \textbf{Bold}: the best result. \underline{Underlined}: OWSM v3.1 outperforms OWSM v3.}
  \label{tab:translation-en-x}
  \vskip -0.1in
  \centering
  \resizebox {0.95\linewidth} {!} {
  \begin{tabular}{cccccc}
    \toprule
    \multirow{2}{*}{Target} & \multirow{2}{*}{Training Data (h)} & OWSM v3 & \multicolumn{3}{c}{OWSM v3.1 (ours)} \\
    \cmidrule(lr){3-3}
    \cmidrule(lr){4-6}
    & & medium & base & small & medium\\
    \midrule
    German & 14.0K & \textbf{25.4} & 14.6 & 22.8 & \textbf{25.4} \\
    Catalan & 0.4K & \textbf{20.0} & 7.7 & 15.9 & 19.6\\
    Chinese & 13.7K & \textbf{33.4} & 14.5 & 26.7 & 32.1\\
    Persian & 0.8K & 9.5 & 3.0 & 7.7 &\underline{\textbf{10.1}}\\
    Estonian & 0.4K & \textbf{7.8} & 1.8 & 5.8 & 7.7\\
    Mongolian & 0.4K & 3.1  & 1.0 & 3.3 &\underline{\textbf{4.6}}\\
    Turkish & 0.9K & 6.1  & 1.2 & 4.8 &\underline{\textbf{6.5}}\\
    Arabic & 0.9K & 6.6  & 1.6 & 5.1 &\underline{\textbf{7.2}}\\
    Swedish & 0.4K & 19.9 & 8.1 & 16.6 &\underline{\textbf{20.3}}\\
    Latvian & 0.4K & 6.3  & 1.3 & 4.4 &\underline{\textbf{6.4}}\\
    Slovenian & 0.4K & 8.6 & 0.7 & 5.7 &\underline{\textbf{9.0}}\\
    Tamil & 0.4K & 0.0 & 0.0 & 0.0 &0.0\\
    Japanese & 1.0K & 17.3 & 8.7 & 16.4 &\underline{\textbf{19.6}}\\
    Indonesian & 0.4K & 14.5 & 5.1 & 12.4 &\underline{\textbf{16.1}}\\
    Welsh & 0.4K & \textbf{15.9} & 4.5 & 11.6 & 15.3\\
    \midrule
    \multicolumn{2}{l}{Ave. BLEU ($\uparrow$)} & 13.0 & 4.9 & 10.6 & \underline{\textbf{13.3}}\\
    \multicolumn{2}{l}{Speed-up ($\uparrow$)} & 1.00x & \textbf{3.00x} & 2.43x & \underline{1.25x}\\
    \bottomrule
  \end{tabular}
  }
  \vskip -0.1in
\end{table}
\endgroup

\subsection{Speech translation}
\label{subsec:st}

We evaluate ST on CoVoST-2 test sets~\cite{covost2}. For English-to-X, we utilize all 15 directions. For X-to-English, we report the results of directions where OWSM has more than 100 hours of training data. For other directions with very limited training data like Japanese- or Chinese-to-English, OWSM usually does not work~\cite{asru23-owsm}. We also record the average decoding time of each test set on an NVIDIA A40 GPU and calculate the relative decoding speed compared to OWSM v3.

For X-to-English (shown in \autoref{tab:translation-x-en}), the proposed OWSM v3.1 medium achieves consistently higher BLEU scores than OWSM v3. The average BLEU is improved from 18.8 to 20.1. OWSM v3.1 is also 16\% faster than OWSM v3 during inference. Compared to Whisper, OWSM v3.1 performs still worse due to limited training data. But OWSM v3.1 has a faster inference speed than Whisper at each scale, thanks to the larger time shift in the encoder (40 ms vs. 20 ms) and the smaller decoder.

For English-to-X (shown in \autoref{tab:translation-en-x}), OWSM v3.1 outperforms OWSM v3 in 9 of 15 directions. The average BLEU is slightly improved from 13.0 to 13.3 and the inference speed is 25\% faster. Note that Whisper cannot perform translation in these directions.

\begingroup
\begin{table}[tb]
  \caption{WER ($\downarrow$) of long-form ASR on TEDLIUM. \textbf{Bold}: the best result. \underline{Underlined}: OWSM v3.1 outperforms OWSM v3.}
  \label{tab:longform-asr}
  \vskip -0.12in
  \centering
  \resizebox {\linewidth} {!} {
  \begin{tabular}{ccccccc}
    \toprule
    \multicolumn{3}{c}{Whisper} & OWSM v3 & \multicolumn{3}{c}{OWSM v3.1 (ours)}\\
    \cmidrule(lr){1-3}
    \cmidrule(lr){4-4}
    \cmidrule(lr){5-7}
    base & small & medium & medium & base & small & medium \\
    \midrule
    5.3 & 4.4 & \textbf{3.8} & 9.2 & 9.6 & 6.7 & \underline{5.7}\\
    \bottomrule
  \end{tabular}
  }
  \vskip -0.2in
\end{table}
\endgroup

\begingroup
\begin{table}[t]
  \caption{Accuracy \% ($\uparrow$) of LID on FLEURS~\cite{FLEURS}.}
  \label{tab:lid}
  \vskip -0.1in
  \centering
  \resizebox {\linewidth} {!} {
  \begin{tabular}{ccccccc}
    \toprule
    \multicolumn{3}{c}{Whisper} & OWSM v3 & \multicolumn{3}{c}{OWSM v3.1 (ours)}\\
    \cmidrule(lr){1-3}
    \cmidrule(lr){4-4}
    \cmidrule(lr){5-7}
    base & small & medium & medium & base & small & medium \\
    \midrule
    47.6 & 53.1 & 54.8 & \textbf{81.4} & 41.9 & 67.1 & 75.6\\
    \bottomrule
  \end{tabular}
  }
  \vskip -0.1in
\end{table}
\endgroup

\begingroup
\setlength{\tabcolsep}{3pt}
\begin{table}[t]
  \caption{F1 scores ($\uparrow$) of SLU tasks on SLUE-PERB~\cite{slue-perb}.}
  \label{tab:slueperb}
  \vskip -0.1in
  \centering
  \resizebox {\linewidth} {!} {
  \begin{tabular}{cccc}
    \toprule
    Task & Metric & OWSM v3 & OWSM v3.1 (ours) \\
    \midrule
    Sentiment Analysis & F1 score & \textbf{60.1} & 56.2\\
    Named Entity Recognition & F1 score & 54.8 & \textbf{65.8}\\
    Named Entity Localization & frame-F1 & 40.5 & \textbf{50.4}\\
    Dialogue Act Classification & F1 score & 56.5 & \textbf{64.8}\\
    \bottomrule
  \end{tabular}
  }
  \vskip -0.15in
\end{table}
\endgroup

\begingroup
\setlength{\tabcolsep}{3pt}
\begin{table}[t]
  \caption{WER ($\downarrow$) of zero-shot contextual biasing.}
  \label{tab:biasing}
  \vskip -0.1in
  \centering
  \resizebox {0.95\linewidth} {!} {
  \begin{tabular}{lcccccc}
    \toprule
    \multirow{2}{*}{OWSM v3.1} & \multicolumn{3}{c}{LibriSpeech test-clean} & \multicolumn{3}{c}{LibriSpeech test-other}\\
    \cmidrule(lr){2-4}
    \cmidrule(lr){5-7}
     & WER & U-WER & B-WER & WER & U-WER & B-WER \\
    \midrule
    base & 3.88 & 2.45 & 15.47 & 9.48 & 6.89 & 32.17\\
    ~~ + biasing & 4.37 & 3.09 & 14.79 & 12.49 & 10.45 & 30.36\\
    \midrule
    small & 2.68 & 1.63 & 11.27 & 6.16 & 4.21 & 23.27\\
    ~~ + biasing & 2.58 & 1.75 & 9.32 & 5.89 & 4.48 & 18.34\\
    \midrule
    medium & 2.59 & 1.61 & 10.61 & 5.31 & 3.52 & 21.12\\
    ~~ + biasing & 2.24 & 1.62 & 7.31 & 5.03 & 3.86 & 15.35\\
    \bottomrule
  \end{tabular}
  }
  \vskip -0.15in
\end{table}
\endgroup

\subsection{Long-form speech recognition}

\autoref{tab:longform-asr} presents long-form English ASR results on the TEDLIUM test set~\cite{tedlium3}. Similar to~\cite{whisper, asru23-owsm}, OWSM takes an entire audio recording as input and generates transcripts in chunks. Each chunk has a fixed length of 30s and is gradually shifted based on the predicted timestamps. The proposed OWSM v3.1 medium achieves a WER of 5.7\%, compared to 9.2\% of OWSM v3. 
This demonstrates the robustness of OWSM v3.1 against long-form audio; the predicted timestamps might also be more accurate. 
OWSM v3.1 still falls behind Whisper, likely because (1) our training data is only around a quarter of Whisper's training data, and (2) many public datasets used by OWSM do not provide unsegmented long-form data and we have to use the segmented short audio for training, which leads to a mismatch between training and inference. In the future, we will add more long-form data to mitigate this issue.

\subsection{Language identification}

\autoref{tab:lid} shows the accuracy of language identification on the FLEURS test set. We notice a degradation of OWSM v3.1 compared to the previous OWSM v3, but OWSM v3.1 medium is still much better than Whisper medium because our model uses the massively multilingual FLEURS and CommonVoice data for training. We also find that OWSM v3.1 benefits more from scaling up compared to Whisper. From base to medium, the accuracy of OWSM v3.1 is almost doubled (41.9\% to 75.6\%), while the accuracy of Whisper is only slightly increased (47.6\% to 54.8\%). A possible reason is that OWSM supports more languages for ASR and language pairs for ST, which is more challenging for smaller models to learn.

\subsection{Spoken language understanding via fine-tuning}

Pre-trained speech models can be applied to downstream tasks via fine-tuning, which generally improves performance~\cite{pretraining-for-slu}. We take spoken language understanding (SLU) as an example and evaluate OWSM on the recently proposed SLUE-PERB benchmark~\cite{slue-perb}.
Specifically, the pre-trained speech encoder is frozen and a randomly initialized shallow decoder is trained on task-specific SLU data. The model is then evaluated on the corresponding SLU test data. This evaluation procedure is similar to the widely used SUPERB benchmark~\cite{superb}. We consider four SLU tasks, i.e., sentiment analysis (SA), named entity recognition (NER), named entity localization (NEL), and dialog act classification (DAC). As shown in \autoref{tab:slueperb}, the proposed OWSM v3.1 medium outperforms the previous v3 model by a large margin in NER, NEL, and DAC, confirming the strong capacity of our E-Branchformer encoder.

\subsection{Emergent ability for zero-shot contextual biasing}

OWSM generates ASR or ST hypotheses conditioned on an optional text prompt.
During training, the previous sentence in the same recording is used as a prompt according to the probability of 0.5. During inference, the user can provide a prompt to potentially adjust the output.
An application of this feature is zero-shot contextual biasing, which aims to improve the ASR performance of rare words by providing a list of biasing words containing true targets and many distractions~\cite{deep-biasing}. We evaluate OWSM v3.1 models on the LibriSpeech biasing test sets created by~\cite{deep-biasing}. Specifically, we use 100 biasing words separated by spaces as the prompt and perform greedy decoding. Unlike Section~\ref{subsec:en-asr}, we do not use any text normalizer to match the condition in~\cite{deep-biasing}. Contextual biasing aims to reduce the biased WER (B-WER) while maintaining the unbiased WER (U-WER).
\autoref{tab:biasing} shows the WERs of our three models. Compared to ASR without biasing, the base model shows minor improvements on B-WER but much larger degradations on U-WER, indicating that it cannot distinguish between useful contextual information and distractions. In contrast, small and medium models greatly reduce B-WER and mostly maintain U-WER, demonstrating that these models can extract and utilize useful contextual information in a zero-shot manner.
The phenomenon that the smaller OWSM performs very poorly in zero-shot biasing ASR while larger ones perform well reveals that \textbf{speech foundation models also have the emergent ability}, which has been widely observed in LLMs~\cite{emergent-ability}.

\section{Conclusion and future work}

We present OWSM v3.1, a family of Open Whisper-style Speech Models based on E-Branchformer, ranging from 100M to 1B parameters. Although trained on the same amount of data, OWSM v3.1 achieves better results than the previous OWSM v3 in the vast majority of evaluation sets, while showing up to 25\% faster inference speeds. We further investigate the emergent ability of speech foundation models using zero-shot contextual biasing ASR, which verifies the benefit of scaling up. To extend the accessibility of our model, we provide a model trained on a subset of data with low license restrictions.
We will publicly release the code, pre-trained model weights, and training logs to promote transparency and facilitate the development of foundation models in the speech field.

A limitation is that this work does not enhance the quantity or quality of training data, which might lead to suboptimal performance in low-resource languages.
Future research directions include exploring the impact of data diversity on model performance, adding more public data like YODAS~\cite{yodas} for better performance, compressing the pre-trained model for better efficiency~\cite{distilhubert, parp, pruning, dphubert, i3d-dynamic, distil-whisper}, and exploring various downstream applications such as SLU~\cite{pretraining-for-slu, arora2023universlu} and speech language models~\cite{slm, tang2023salmonn}.

\section{Acknowledgements}
We would like to thank Amazon AGI for funding.
We use PSC Bridges2 and NCSA Delta via ACCESS CIS210014, by National Science Foundation grants \#2138259, \#2138286, \#2138307, \#2137603, and \#2138296.

\section{References}
\printbibliography

\end{document}